\documentclass{article}

\usepackage{arxiv}

\usepackage[utf8]{inputenc} 
\usepackage[T1]{fontenc}    
\usepackage{hyperref}       
\usepackage{url}            
\usepackage{booktabs}       
\usepackage{amsfonts}       
\usepackage{nicefrac}       
\usepackage{microtype}      
\usepackage{lipsum}
\usepackage{graphicx}
\graphicspath{ {./images/} }
\usepackage{caption}  
\usepackage{amsmath}
\usepackage{multirow}

\usepackage{listings}
\NewDocumentCommand{\codeword}{v}{%
\texttt{{#1}}%
}
\providecommand{\aboveskip}{10pt}
\providecommand{\belowskip}{10pt}

\title{The impact of model size on catastrophic forgetting in Online Continual Learning}

\author{
 Eunhae Lee \\
  Department of Electrical Engineering and Computer Science\\
  Massachusetts Institute of Technology\\
  Cambridge, MA 02139 \\
  \texttt{eunhae@mit.edu} \\
}

\begin{document}
\maketitle
\begin{abstract}
This study investigates the impact of model size on Online Continual Learning performance, with a focus on catastrophic forgetting. Employing ResNet architectures of varying sizes, the research examines how network depth and width affect model performance in class-incremental learning using the SplitCIFAR-10 dataset. Key findings reveal that larger models do not guarantee better Continual Learning performance; in fact, they often struggle more in adapting to new tasks, particularly in online settings. These results challenge the notion that larger models inherently mitigate catastrophic forgetting, highlighting the nuanced relationship between model size and Continual Learning efficacy. This study contributes to a deeper understanding of model scalability and its practical implications in Continual Learning scenarios.
\end{abstract}


\section{Introduction}
One of the biggest unsolved challenges in Continual Learning (CL) is the prevention of forgetting previously learned information upon acquiring new information. Known as “catastrophic forgetting,” this phenomenon is particularly pertinent in scenarios where AI systems must adapt to new data without losing valuable insights from past experiences \cite{McCloskey1989, goodfellow2015empirical, kirkpatrick2017overcoming}. Numerous studies have investigated different approaches to solving this problem in the past years, mostly around proposing innovative strategies to modify the way models are trained and measuring its impact on model performance, such as accuracy and forgetting.

Yet, compared to the numerous amount of studies done in establishing new strategies and evaluative approaches in visual continual learning, there is surprisingly little discussion on the impact of model size on CL performance. It is commonly known that the size of a deep learning model (the number of parameters) is known to play a crucial role in its learning capabilities 
 \cite{hu2021model, Bianco_2018}. Given the limitations in computational resources in most real-world circumstances, it is often not practical or feasible to choose the largest model available. In addition, sometimes smaller models perform just as well as larger models in specific contexts \cite{Bressem_2020}. Given this context, a better understanding of how model size impacts performance in a Continual Learning setting can provide insights and implications for real-world deployment of CL systems.

This research examines how network depth and width affect model performance in class-incremental learning in both online and offline Continual Learning settings, using ResNet architectures of varying depths and widths. The hypothesis is set forth based on existing literature on model size and performance and is tested through an empirical experiment using ResNet models trained from scratch. The study aims to shed light on whether larger models truly offer an advantage in mitigating catastrophic forgetting, or if the reality is more nuanced.

\section{Related Work}
\label{sec:relatedwork}
\subsection{Online Continual Learning}
Continual Learning (CL), also known as Lifelong Learning or Incremental Learning, is an approach that seeks to continually learn from non-iid data streams without forgetting previously acquired knowledge. The challenge in Continual Learning is generally known as the stability-plasticity dilemma \cite{mermillod2013-dilemma}, and the goal of Continual Learning is to strike a balance between learning stability and plasticity.

Traditional CL models assume new data arrives task by task, each with stable data distribution, enabling offline training. However, this requires having access to all task data, which can be impractical due to privacy or resource limitations. This study will focus on a more realistic setting of Online Continual Learning (OCL), where data arrives in smaller batches and are not accessible after training, requiring models to learn from a single pass over an online data stream \cite{soutif-cormerais_comprehensive_2023, cai_online_2021, mai_online_2021}. This allows the model to learn data in real time.

Online Continual Learning can involve adapting to new classes (class-incremental) or changing data characteristics (domain-incremental). For class-incremental learning, the goal is to continually expand the model's ability to recognize an increasing number of classes, maintaining its performance on all classes it has seen so far, despite not having continued access to the old class data \cite{soutif-cormerais_comprehensive_2023, ghunaim_real-time_2023}. More recently, there have been studies investigating Unsupervised Continual Learning \cite{yu_scale_2023, madaan_representational_2022}. However, to narrow the scope of the vast CL landscape to focus on learning the impact of model size on CL performance, this study will focus on the more common problem of class-incremental learning in supervised image classification in this study.

\subsection{Continual Learning techniques}
\label{sec:cl-techniques}
Popular strategies to mitigate catastrophic forgetting in Continual Learning generally fall into three buckets \cite{ghunaim_real-time_2023}:

\begin{enumerate}
    \item \textbf{Regularization-based approaches.} Regularization-based approaches modify the classification objective to preserve past representations or foster more insightful representations, such as Elastic Weight Consolidation (EWC) \cite{kirkpatrick2017overcoming} and Learning without Forgetting (LwF) \cite{li_learning_2017}.
    \item \textbf{Memory-based approaches.} Memory- or replay-based approaches replay samples retrieved from a memory buffer along with every incoming mini-batch, including Experience Replay (ER) \cite{chaudhry2019tiny} and Maximally Interfered Retrieval \cite{aljundi2019online}, with variations on how the memory is retrieved and how the model and memory are updated.
    \item \textbf{Architectural approaches.} Architectural approaches include parameter-isolation approaches where new parameters are added for new tasks and leaving previous parameters unchanged such as Progressive Neural Networks (PNNs) \cite{rusu2022progressive}. 
\end{enumerate}

Many methods also combine two or more of these approaches, such as Averaged Gradient Episodic Memory (A-GEM) \cite{chaudhry2019efficient}  and Incremental Classifier and Representation Learning (iCaRL) \cite{rebuffi2017icarl}.

Specifically, this study will focus on \textbf{Experience Replay (ER)}, a classic replay-based method widely used for Online Continual Learning. Despite its simplicity, recent studies have shown ER still outperforms many of the newer methods that have come after that, especially for Online Continual Learning \cite{soutif-cormerais_comprehensive_2023, mai_online_2021, ghunaim_real-time_2023}.

\subsection{Model size and performance}
\label{sec:modelsizeandperf}
It is generally known across literature that larger and deeper models lead to increased performance \cite{hu2021model}. Bianco et al. surveyed key performance-related metrics to compare across various architectures, including accuracy, model complexity, computational complexity, and accuracy density \cite{Bianco_2018}. The relationship between model width and performance is also been discussed \cite{hu2021model}, albeit less frequently.

In 2015, Residual Networks (ResNets) were introduced by He et al.\cite{he2015deep}, which was a major innovation in computer vision. ResNets tackled the problem of degradation that occurs in deep networks through the use of residual blocks to increase the accuracy of deeper models. Residual blocks that contain two or more layers are stacked together, and “skip connections” are used in between these blocks. The skip connections act as an alternate shortcut for the gradient to pass through, which alleviates the issue of vanishing gradient. They also make it easier for the model to learn identity functions. As a result, ResNets improve the efficiency of deep neural networks with more neural layers while minimizing the percentage of errors. The authors compare models of different depths (composed of 18, 34, 50, 101, and 152 layers) and show that accuracy increases with the depth of the model \cite{he2015deep}.

\begin{table}[htbp]
  \centering
  \begin{tabular}{lccccc}
    \toprule
    & \textbf{ResNet18} & \textbf{ResNet34} & \textbf{ResNet50} & \textbf{ResNet101} & \textbf{ResNet152} \\
    \midrule
    Number of Layers & 18 & 34 & 50 & 101 & 152 \\
    Number of Parameters & $\sim$11.7 million & $\sim$21.8 million & $\sim$25.6 million & $\sim$44.5 million & $\sim$60 million \\
    Top-1 Accuracy & 69.76\% & 73.31\% & 76.13\% & 77.37\% & 78.31\% \\
    Top-5 Accuracy & 89.08\% & 91.42\% & 92.86\% & 93.68\% & 94.05\% \\
    FLOPs & 1.8 billion & 3.6 billion & 3.8 billion & 7.6 billion & 11.3 billion \\
    \bottomrule
  \end{tabular}
  \caption{Comparison of ResNet Architectures. Accuracy scores are based on the ImageNet benchmark.}  
  \label{tab:table1}
\end{table}

This leads to the question: do larger models perform better in Continual Learning? While much of the focus in CL research has often been on developing various techniques and establishing benchmarks, the impact of model scale remains a less explored path.

Moreover, recent studies on model scales in slightly different contexts have shown conflicting results. Luo et al. \cite{luo2023empirical} highlight a direct correlation between increasing model size and the severity of catastrophic forgetting in large language models (LLMs). They test models of varying sizes from 1 to 7 billion parameters. Yet, Dyer et al. \cite{dyer2022} show a contrasting perspective in the context of pre-trained deep learning models. Their results show that large, pre-trained ResNets and Transformers are more resistant to forgetting than randomly initialized, trained-from-scratch models, and that this tendency increases with the scale of the model and the size of the dataset used for pre-training.

The relative lack of discussion on model size and the conflicting perspectives among existing studies indicate that the answer to the question is far from being definitive. The next two sections will elaborate on the specific approach and parameters of the study.

\section{Method}

\subsection{Problem definition}
Online Continual Learning can be defined as follows \cite{cai_online_2021, ghunaim_real-time_2023}: The objective is to learn a function \( f_\theta : \mathcal{X} \rightarrow \mathcal{Y} \) with parameters \( \theta \) that predicts the label \( Y \in \mathcal{Y} \) of the input \( \mathbf{X} \in \mathcal{X} \). Over time steps \( t \in \lbrace 1, 2, \ldots, \infty \rbrace \), a distribution-varying stream \( \mathcal{S} \) reveals data sequentially, which is different from classical supervised learning.

At every time step,

\begin{enumerate}
    \item \( \mathcal{S} \) reveals a set of data points (images) \( \mathbf{X}_t \sim \pi_t \) from a non-stationary distribution \( \pi_t \)
    \item Learner \( f_\theta \) makes predictions \( \hat{Y}_t \) based on current parameters \( \theta_t \)
    \item \( \mathcal{S} \) reveals true labels \( Y_t \)
    \item Compare the predictions with the true labels, compute the training loss \( L(Y_t, \hat{Y}_t) \)
    \item Learner updates the parameters of the model to \( \theta_{t+1} \)
\end{enumerate}

\subsection{Task-agnostic and boundary-agnostic}
\label{sec:section3.2}
In the context of class-incremental learning, the definitions of task-agnostic and boundary-agnostic from Soutif et al. \cite{soutif-cormerais_comprehensive_2023} are adopted. A \textit{task-agnostic} setting refers to when task labels are not available, which means the model does not know that the samples belong to a certain task. A \textit{boundary-agnostic} setting is considered, where information on task boundaries is unavailable. This means that the model does not know when the data distribution changes to a new task. To reflect a more realistic Continual Learning setting, this study assumes a setting that is both task-agnostic and boundary-agnostic.

\begin{table}[htbp]
  \centering
  \begin{tabular}{@{}lcc@{}}
    \toprule
    & \textbf{Yes} & \textbf{No} \\
    \midrule
    Task labels & Task-aware & Task-agnostic \\
    Task boundaries & Boundary-aware & Boundary-agnostic \\
    \bottomrule
  \end{tabular}
  \caption{Task labels and task boundaries.}
  \label{tab:table2}
\end{table}

\subsection{Experience Replay (ER)}
\label{sec:section3.3}
In a class-incremental learning setting, the nature of the Experience Replay (ER) method aligns well with task-agnostic and boundary-agnostic settings. This is because ER focuses on replaying a subset of past experiences, which helps in maintaining knowledge of previous classes without needing explicit task labels or boundaries. This characteristic of ER allows it to adapt to new classes as they are introduced, while retaining the ability to recognize previously learned classes, making it inherently suitable for task-agnostic and boundary-agnostic Continual Learning scenarios.

In terms of implementation, ER involves randomly initializing an external memory buffer $\mathcal{M}$, then implementing \codeword{before_training_exp}
and \codeword{after_training_exp} callbacks to use the dataloader to create mini-batches with samples from both the training stream and the memory buffer. Each mini-batch is balanced so that all tasks or experiences are equally represented in terms of stored samples \cite{lomonaco2021avalanche}. As ER is known to be well-suited for Online Continual Learning as explained in Section \ref{sec:cl-techniques}, it will be the go-to method used to compare performances across models of varying sizes.

\subsection{Benchmark}
For this study, the SplitCIFAR-10 \cite{lomonaco2021avalanche} is used as the main benchmark to compare the performance across models of different sizes. SplitCIFAR-10 splits the popular CIFAR-10 dataset into five tasks with disjoint classes, each task including two classes. Each task has 10,000 3×32×32 images for training and 2000 images for testing. The model is exposed to these tasks or experiences sequentially, which simulates a real-world scenario where a learning system is exposed to new categories of data over time. This is suitable for class-incremental learning scenarios. This benchmark is used for both testing online and offline Continual Learning in this study.

\subsection{Metrics}
To evaluate the continual learning performance of each model, key metrics that have been established in earlier work in Online Continual Learning are used.

\paragraph{Average Anytime Accuracy (AAA) \cite{caccia_new_2022}} 
The concept of Average Anytime Accuracy serves as an indicator of a model’s overall performance throughout its learning phase, extending the idea of average incremental accuracy to include continuous assessment scenarios. This metric assesses the effectiveness of the model across all stages of training, rather than at a single endpoint, offering a more comprehensive view of its learning trajectory.

\begin{equation}
\text{AAA} = \frac{1}{T} \sum_{t=1}^{T} (\text{AA})_t
\end{equation}

\paragraph{Average Cumulative Forgetting (ACF) \cite{soutif-cormerais_comprehensive_2023, soutifcormerais2021importance}} 
First, \textbf{Cumulative Accuracy} ($b_k^t$) for task $k$ after the model has been trained up to task $t$ is defined. It computes the mean accuracy over the evaluation set $E^k_\Sigma$, which contains all instances $x$ and their true labels $y$ up to task $k$. The model's prediction for each instance is given by $\underset{c \in C^k_\Sigma}{\text{arg max }} f^t(x)_c$, which selects the class $c$ with the highest predicted logit $f^t(x)_c$. The indicator function $1_y(\hat{y})$ outputs 1 if the prediction matches the true label, and 0 otherwise. The sum of these outputs is then averaged over the size of the evaluation set to compute the cumulative accuracy.

\begin{equation}
    b_k^t = \frac{1}{|E^k_\Sigma|} \sum_{(x,y) \in E^k_\Sigma} 1_y(\underset{c \in C^k_\Sigma}{\text{arg max }} f^t(x)_c)
\end{equation}

From Cumulative Accuracy, we can calculate the \textbf{Average Cumulative Forgetting} ($F_{\Sigma}^t$) by setting the cumulative forgetting about a previous cumulative task $k$, then averaging over all tasks learned so far:

\begin{equation}
    F_{\Sigma}^t = \frac{1}{t-1} \sum_{k=1}^{t-1} \max_{i=1,...,t} \left( b_k^i - b_k^t \right)
\end{equation}

\paragraph{Average Accuracy (AA) and Average Forgetting (AF) \cite{mai_online_2021}}
$a_{i,j}$ is the accuracy evaluated on the test set of task $j$ after training the network from task 1 to $i$, while $i$ is the current task being trained. Average Accuracy (AA) is computed by averaging this over the number of tasks.

\begin{equation}
    \text{Average Accuracy} (AA_i) = \frac{1}{i} \sum_{j=1}^{i} a_{i,j}
\end{equation}

Average Forgetting measures how much a model's performance on a previous task (task $j$) decreases after it has learned a new task (task $i$). It is calculated by comparing the highest accuracy the model $\max_{l \in {1, \ldots, k-1}} (a_{l, j})$ had on task $j$ before it learned task $k$, with the accuracy $a_{k, j}$ on task $j$ after learning task $k$.

\begin{equation}
    \text{Average Forgetting}(F_i) = \frac{1}{i - 1} \sum_{j=1}^{i-1} f_{i,j}
\end{equation}

\begin{equation}
    f_{k,j} = \max_{l \in \{1,...,k-1\}} (a_{l,j}) - a_{k,j}, \quad \forall j < k
\end{equation}

In the context of class-incremental learning, the classical concept of  forgetting (Average Forgetting) may not provide meaningful insight due to its tendency to increase as the complexity of the task grows (considering there are more classes within the classification problem). Therefore, Soutif et al. \cite{soutif-cormerais_comprehensive_2023} recommended avoiding relying on classical forgetting as a metric in settings of class-incremental learning, both online and offline settings. For this reason, Average Anytime Accuracy (AAA) and Average Cumulative Forgetting (ACF) are used throughout this experiment, although Average Accuracy (AA) and Average Forgetting (AF) are computed as part of the process.

\subsection{Model selection}
To compare the learning performance across different model depths, popular ResNet architectures, including ResNet18, ResNet34, and ResNet50 are used. As mentioned in Section \ref{sec:modelsizeandperf}, ResNets were designed to increase the performance of deeper neural networks, and their performance metrics are well known. While using custom models with more variability in sizes was a consideration, popular existing architectures were chosen for better reproducibility.

Moreover, in order to observe the effect of model width on CL performance, a slim version of ResNet18 that has been implemented in previous work \cite{lopez-paz_gradient_2017} was used to compare with the performance of ResNet18. The slim version of ResNet18 uses fewer filters per layer, reducing the model width and computational load while keeping the depth of the original model.

While there are more recent versions of ResNet (e.g. ResNeXt \cite{xie2017aggregated}) that have shown to perform better without a significant increase in computational complexity \cite{Bianco_2018}, the original simpler models were chosen for this research to avoid introducing unnecessary variables. ResNet18 and ResNet34 have the basic residual network structure, and ResNet50, ResNet101, and ResNet152 use slightly modified building blocks that have three layers instead of two as in the original residual block structure. This "bottleneck design" was made to reduce training time of larger models. The specifics of the design of these models are detailed in the table from the original paper by He et al. \cite{he2015deep}.


\subsection{Saliency maps}
Saliency maps were utilized in the study to qualitatively visualize the “attention” of the networks of different sizes. Saliency maps are commonly used to understand which areas of the input images are most influential for the model’s predictions. By visualizing the specific areas of an image that a Convolutional Neural Network (CNN) considers important for classification, saliency maps provide insight into the internal representation and decision-making process of the network \cite{simonyan2014deep}.

\section{Experiment}

\subsection{Setup}
The setup of the experiment is as follows:
\begin{itemize}
    \item Each ResNet model was trained from scratch using the Split-CIFAR10 benchmark with 2 classes per task, for 3 epoches with a mini-batch size of 64.
    \item SGD optimizer with a 0.9 momentum and 1e-5 weight decay was used. The initial learning rate is set to 0.01 and the scheduler reduces it by a factor of 0.1 every 30 epochs, as done in \cite{lin_clear_2022}.
    \item Cross entropy loss is used as the criterion, as is common for image classification in Continual Learning.
    \item Basic data augmentation is done on the training data to enhance model robustness and generalization by artificially expanding the dataset with varied, modified versions of the original images.
    \item Each model is trained in both online and offline CL settings, the latter serving as baselines for performance comparison.
    \item Memory size of 500 (representing 1\% of the training dataset) is used to implement Experience Replay (ER).
\end{itemize}

\subsection{Implementation}
The Continual Learning benchmark was implemented using the Avalanche framework \cite{lomonaco2021avalanche}, an open source Continual Learning library, and an adapted version of the code for Online Continual Learning by Soutif et al. \cite{soutif-cormerais_comprehensive_2023}. The experiments were run on NVIDIA Tesla T4 GPU.


\section{Results}
\begin{figure}[h]
    \centering
    \begin{minipage}{0.52\textwidth}
        \centering
        \includegraphics[width=0.96\textwidth]{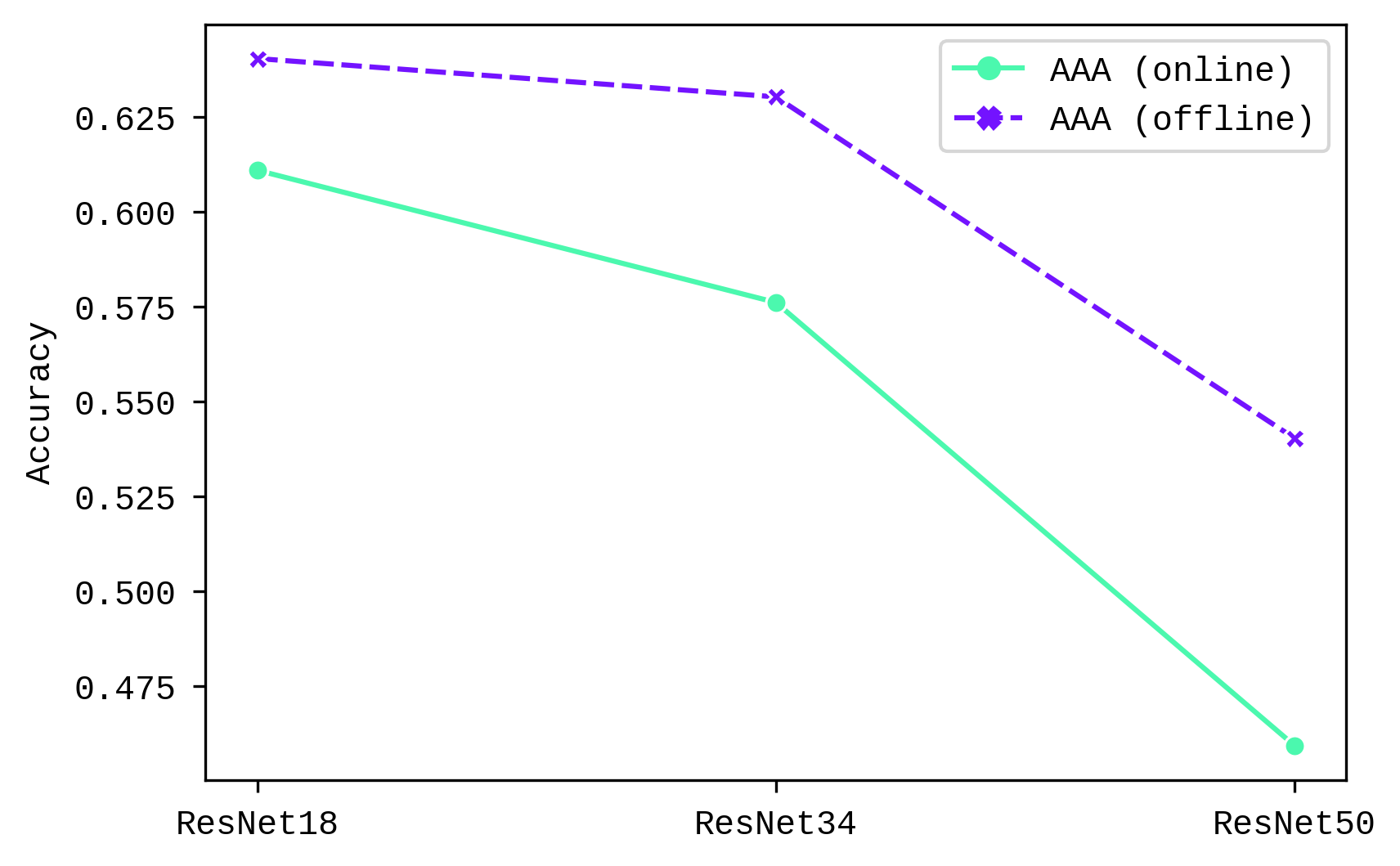} 
        \caption{Average Anytime Accuracy (AAA) of different sized ResNets in Online and Offline Continual Learning}
        \label{fig:accuracy1}
    \end{minipage}\hfill
    \begin{minipage}{0.48\textwidth}
        \centering
        \includegraphics[width=0.96\textwidth]{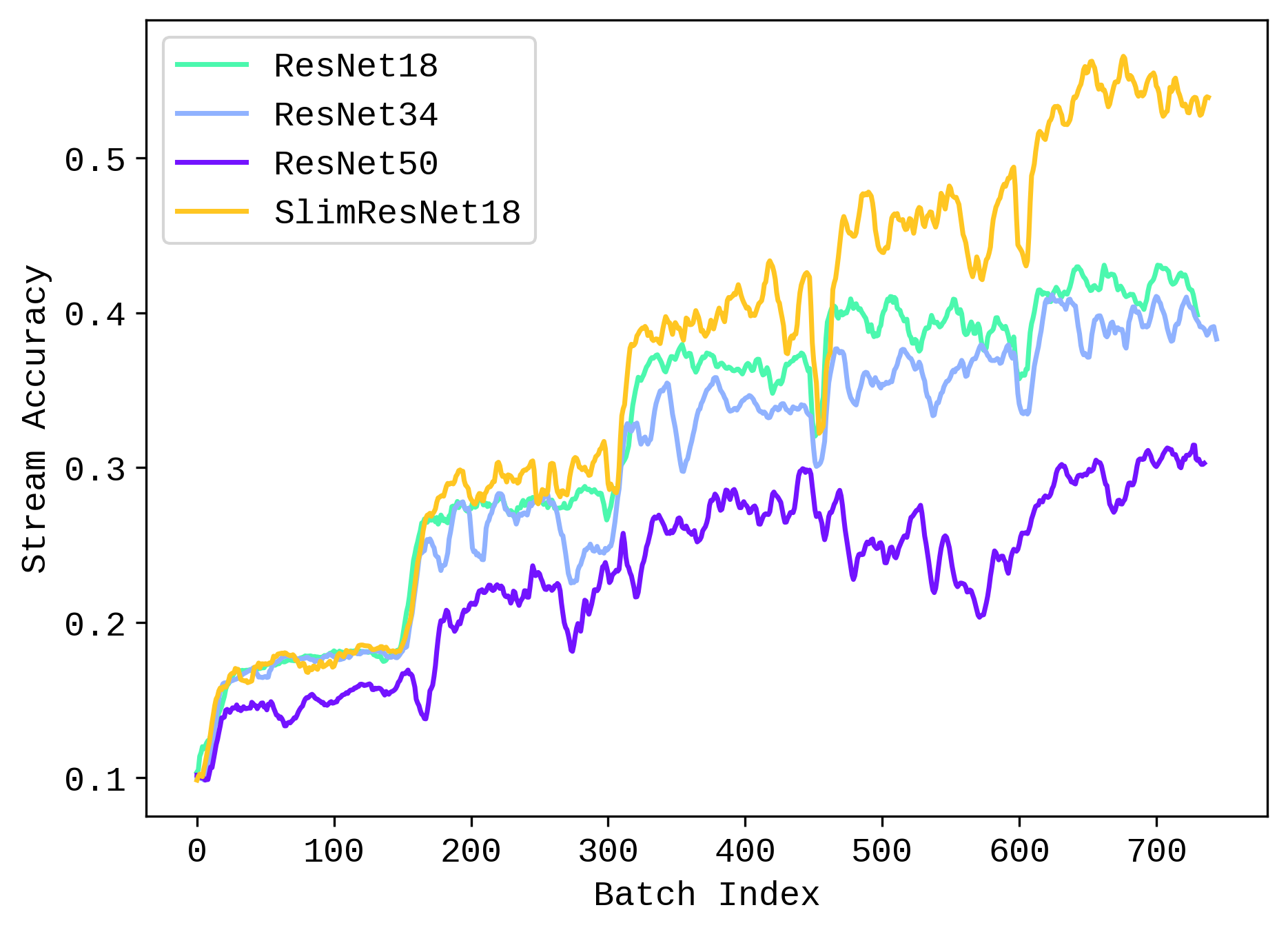} 
        \caption{Validation stream accuracy (Online CL)}
        \label{fig:accuracy2}
    \end{minipage}
\end{figure}

\paragraph{Accuracy decreases as model size increases.}
Average Anytime Accuracy (AAA) decreases for larger models, with a sharper drop in performance between ResNet34 and ResNet50. The decrease in AAA is more significant in online learning than offline learning (Figure \ref{fig:accuracy1}).

\paragraph{Accuracy of larger models grow at a slower pace when training.}
The rate to which validation stream accuracy increases with each task degrade with larger models (Figure \ref{fig:accuracy2}). Interestingly, Slim-ResNet18 shows the highest accuracy and growth trend, suggesting the potential impact of model width on CL performance. This could indicate that larger models are worse at generalizing to a class-incremental learning scenario.

\begin{table}[htbp]
  \centering
  \begin{tabular}{@{}lcc@{}}
    \toprule
    \textbf{Model} & \textbf{Average Anytime Acc (AAA)} & \textbf{Final Average Acc} \\
    \midrule
    Slim-ResNet18 & 0.6645 & 0.5364 \\
    ResNet18 & 0.6110 & 0.3712 \\
    ResNet34 & 0.5761 & 0.3568 \\
    ResNet50 & 0.4594 & 0.3036 \\
    \bottomrule
  \end{tabular}
  \caption{Accuracy metrics across differently sized models (Online CL)}
  \label{tab:resnet_comparison}
\end{table}

\begin{figure}[h]
    \centering
    \begin{minipage}{0.5\textwidth}
        \centering
        \includegraphics[width=0.96\textwidth]{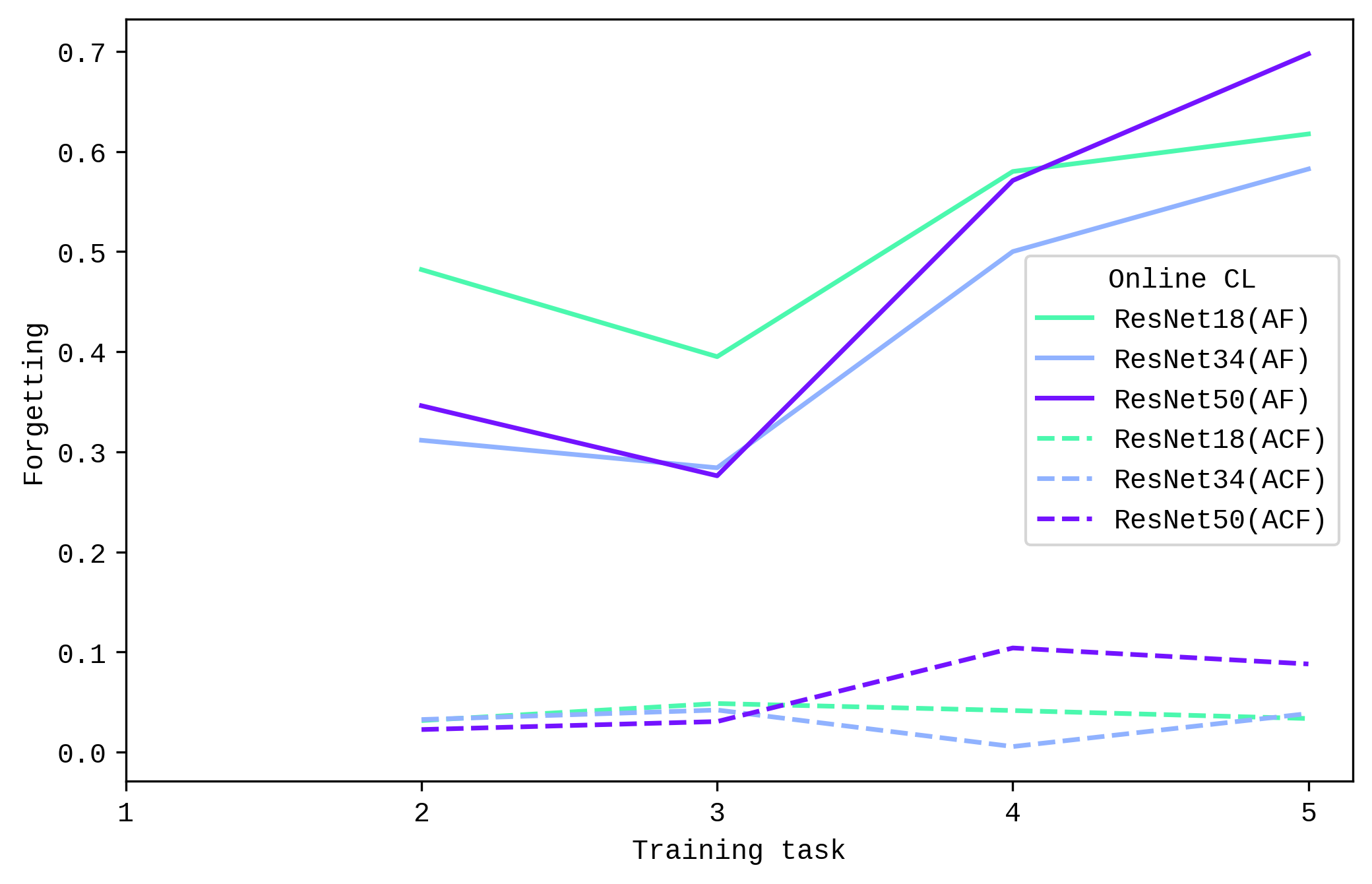} 
    \end{minipage}\hfill
    \begin{minipage}{0.5\textwidth}
        \centering
        \includegraphics[width=0.96\textwidth]{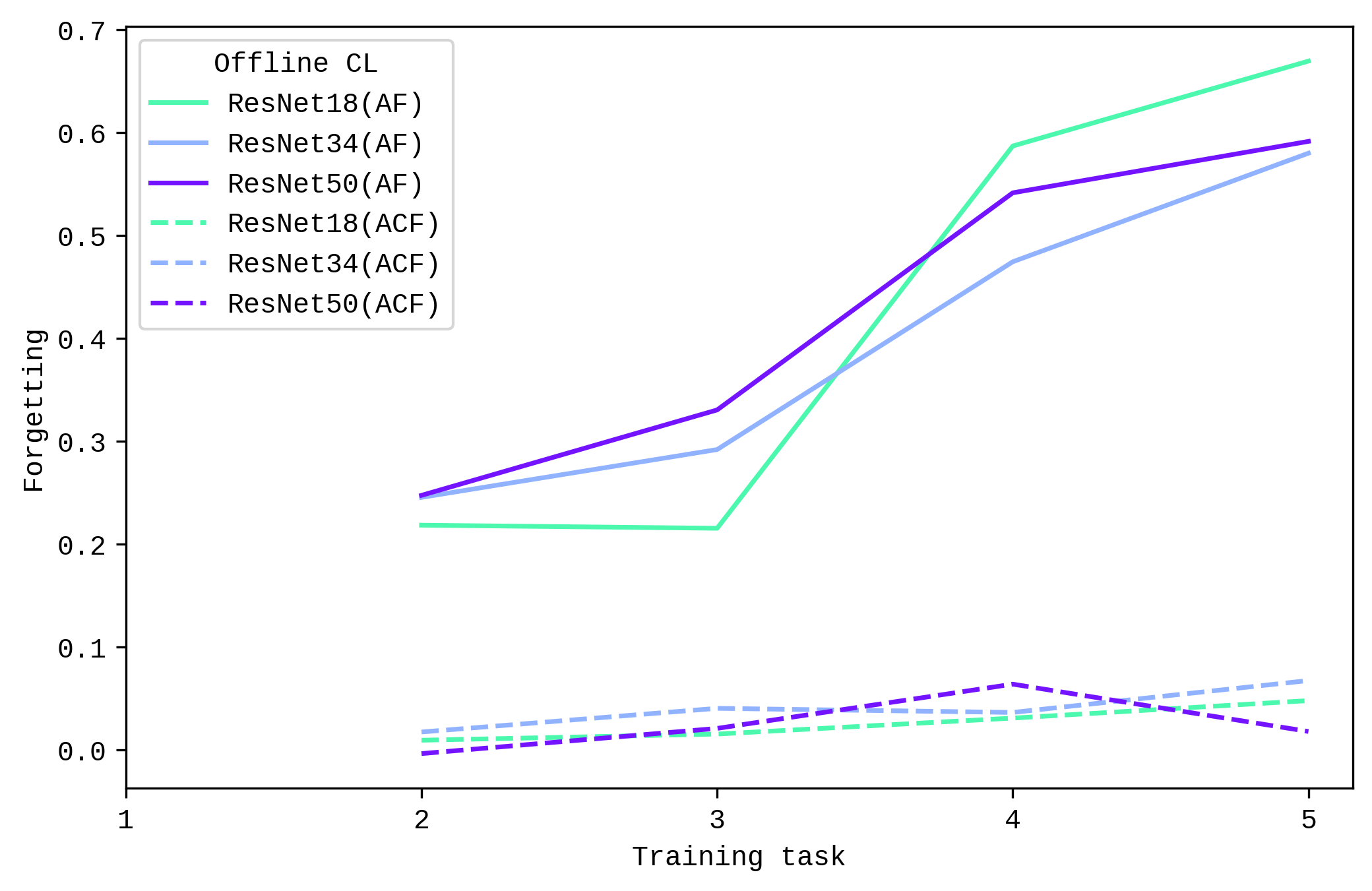} 
    \end{minipage}
    \caption{Forgetting curves, Online CL (left) and Offline CL (right). Solid lines: Average Forgetting (AF); Dotted lines: Average Cumulative Forgetting (ACF)}
    \label{fig:forgetting}
\end{figure}

\paragraph{Forgetting levels show more nuanced results across Online and Offline CL}
In the Online CL setting, the Average Cumulative Forgetting (ACF) is lowest for ResNet34 (with a slight overlap with ResNet18 at Task 5), and highest for ResNet50. A noticeable observation in both ACF and AF is that ResNet50 performed better initially but forgetting levels started to increase after a few tasks. The results for Offline CL setting are slightly different, with ResNet50 having the lowest Average Cumulative Forgetting (ACF) (although with a slight increase at Task 4), followed by ResNet18, and finally ResNet34 (Figure \ref{fig:forgetting}). 

The differences in forgetting between Online and Offline CL settings are aligned with the accuracy metrics in Figure 1, where the performance of ResNet50 decreased more starkly in the Online CL setting.

\begin{figure}[h]
    \includegraphics[width=11cm]{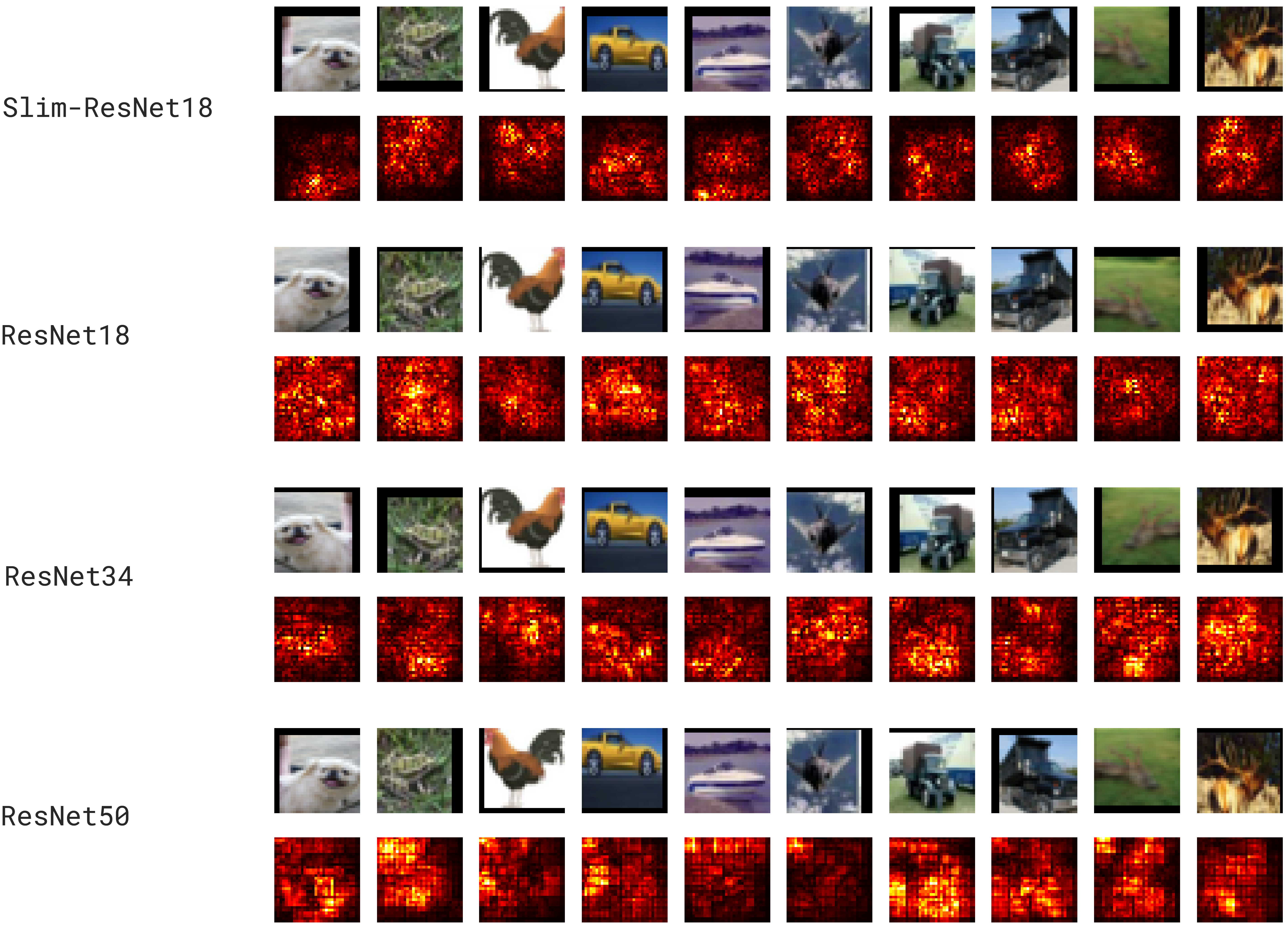}
    \centering
    \caption{Saliency map visualizations for Online CL}
    \label{fig:saliency_on}
\end{figure}

\paragraph{Observation of saliency maps.}
Visual inspection of the saliency maps revealed some interesting observations. When it comes to the ability to highlight intuitive areas of interest in the images, there seemed to be a noticeable improvement from ResNet18 to ResNet34, but this was not necessarily the case from ResNet34 to ResNet50. This phenomenon was more salient in the online CL setting (Figure \ref{fig:saliency_on}).

Interestingly, Slim-ResNet18 seems to be doing better than most of them, certainly better than its plain counterpart ResNet18. A further exploration of model width on performance and representation quality would be an interesting avenue of research.

\section{Discussion}

This research examined the impact of model size on Online Continual Learning performance by comparing key accuracy and forgetting metrics across ResNet models of different depths and widths. These results show that larger models do not necessarily lead to better continual learning performance. We saw that Average Anytime Accuracy (AAA) and stream accuracy dropped progressively with model size, hinting that larger models struggle to generalize to newly trained tasks, especially in an online CL setting. Forgetting curves showed similar trends but with more nuance; larger models perform well at first but suffer from increased forgetting with more incoming tasks. Interestingly, the problem was not as pronounced in the offline CL setting, which highlights the challenges of training models in a more realistic, Online Continual Learning context. Moreover, a qualitative inspection of the saliency maps suggests that the quality of the models’ internal representation also tends to drop with larger models, the difference being the most visible with a change in model width (i.e. ResNet18 vs Slim-ResNet18).

Why do larger models perform worse at Continual Learning? One reason could be that larger models have more parameters, which may make it harder to maintain stability in the learned features as new data is introduced. This makes them more prone to overfitting and forgetting previously learned information, reducing their ability to generalize.

Building on this work, future research could investigate the impact of model size on CL performance by exploring the following questions: (1) Do pre-trained models generalize better in Online Continual Learning settings compared to models trained from scratch? (2) Does longer training improve the relative performance of larger models in CL settings? (3) Can different CL strategies (other than Experience Replay) mitigate the degradation of performance in larger models? (4) Do 
"slimmer" versions of existing models always perform better? (5) How might different hyperparameters (i.e. learning rate) impact CL performance of larger models?

\section{Conclusion}

The results from this study strongly suggests that model size matters when it comes to Continual Learning and forgetting, albeit in nuanced ways. By empirically exploring the under-explored topic of the role of model size on CL performance, these findings contribute to the ongoing discussions on the role of the scale of deep learning models on performance and have implications for future areas of research.

\bibliographystyle{plain}  

\end{document}